\documentclass[sigconf]{acmart}
\AtBeginDocument{%
  \providecommand\BibTeX{{%
    \normalfont B\kern-0.5em{\scshape i\kern-0.25em b}\kern-0.8em\TeX}}}

\setcopyright{acmlicensed}
\copyrightyear{2018}
\acmYear{2018}
\acmDOI{XXXXXXX.XXXXXXX}

\acmConference[In2Writing 2024]{Make sure to enter the correct
  conference title from your rights confirmation emai}{May 11,
  2024}{Honolulu, Hawii}
\acmBooktitle{Woodstock '18: ACM Symposium on Neural Gaze Detection,
 June 03--05, 2018, Woodstock, NY} 
\acmISBN{978-1-4503-XXXX-X/18/06}

\usepackage{xspace}
\usepackage{booktabs}
\usepackage{dirtytalk}
\usepackage{cleveref}
\usepackage{multirow}
\usepackage[inline]{enumitem}
\usepackage{ifthen}
\usepackage{multirow}
\usepackage{url}
\usepackage{makecell}
\usepackage{subcaption}

\title{Generating Educational Materials with Different Levels of Readability using LLMs}

\definecolor{ao(english)}{rgb}{0.0, 0.5, 0.0}

\newcommand{\overview}[1]{}
\newcommand{\cy}[1]{}
\newcommand{\avon}[1]{}
\newcommand{\kenneth}[1]{}
\newcommand{\sanjana}[1]{}
\newcommand{\smb}[1]{}

\newcommand{\ie}{{\it i.e.}}

\newcommand{\llama}{LLaMA-2 70B\xspace}
\newcommand{\mixtral}{Mixtral 8x7B\xspace}
\newcommand{\chatgpt}{GPT-3.5\xspace}

\newenvironment{prompt}%
  {\list{}{\leftmargin=0.2in\rightmargin=0.12in}\item[]}%
  {\endlist}

\setcopyright{none}

\renewcommand\footnotetextcopyrightpermission[1]{}

\begin{document}


\author{Chieh-Yang Huang}
\email{cyhuang@lexile.com}
\affiliation{
    \institution{MetaMetrics Inc.}
    \city{Durham}
    \state{North Carolina}
    \country{USA}
}

\author{Jing Wei}
\email{jwei@lexile.com}
\affiliation{
    \institution{MetaMetrics Inc.}
    \city{Durham}
    \state{North Carolina}
    \country{USA}
}

\author{Ting-Hao Kenneth Huang}
\email{txh710@psu.edu}
\affiliation{
    \institution{Pennsylvania State University}
    \city{University Park}
    \state{Pennsylvania}
    \country{USA}
}

\renewcommand{\shortauthors}{Chieh-Yang Huang, Jing Wei, and Ting-Hao Kenneth Huang}

\begin{abstract}
This study introduces the leveled-text generation task,
aiming to rewrite educational materials to specific readability levels while preserving meaning.
We assess the capability of \chatgpt, \llama, and \mixtral, to generate content at various readability levels through zero-shot and few-shot prompting.
Evaluating 100 processed educational materials reveals that few-shot prompting
significantly improves performance in readability manipulation and information preservation.
\llama performs better in achieving the desired difficulty range, while \chatgpt maintains original meaning.
However, manual inspection highlights concerns such as misinformation introduction and inconsistent edit distribution.
These findings emphasize the need for further research to ensure the quality of generated educational content.


\end{abstract}




\begin{teaserfigure}
  \includegraphics[width=\textwidth]{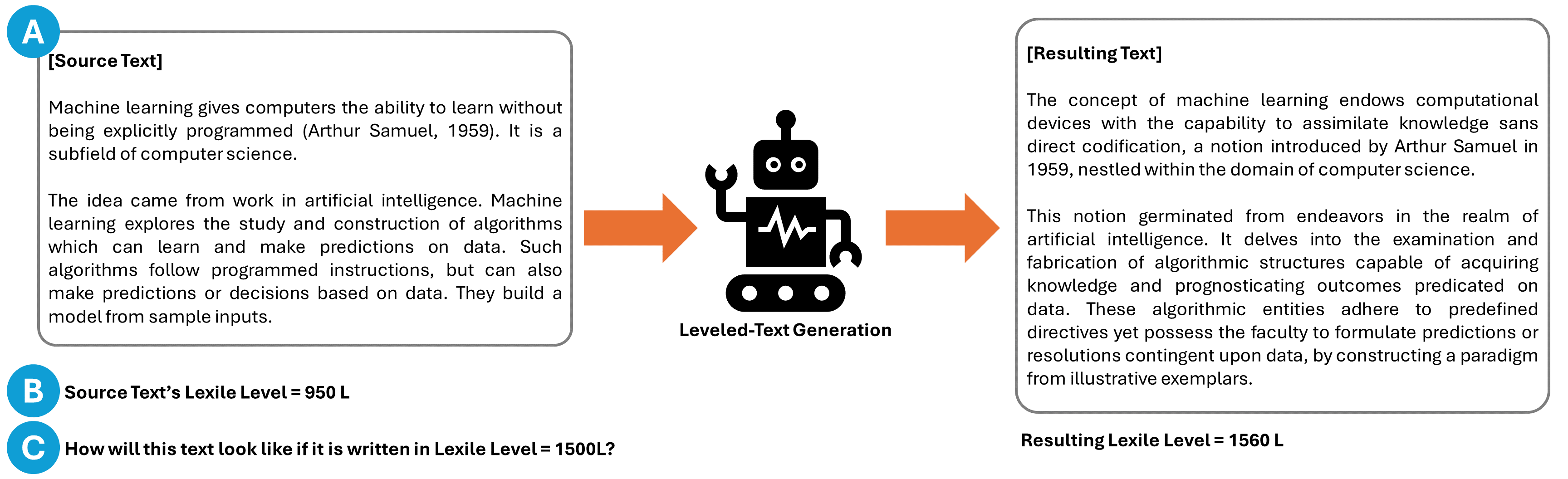}
  \caption{Given (A) a source text, (B) a readability level for the source text, and (C) an intended readability level, the \textbf{leveled-text generation} aims at rewriting the source text in the intended readability level while preserving the meaning. Note that in this paper, we use a widely adopted readability level in the education context, the Lexile Scale. In the provided example, the sentences are longer and the words are less common in the resulting text.}
  \label{fig:task}
\end{teaserfigure}


\maketitle

\section{Introduction}
Prior research has shown that students learn more effectively from reading materials that match their level of readability,
optimally balancing improvement and cognitive load~\cite{lexilereport,yang2021text,doi:10.1080/2331186X.2019.1615740}.
Consequently, curating educational content to meet the diverse reading abilities of students is a crucial step
toward better learning outcomes.
Platforms such as Newsela~\cite{newsela}, Simple English Wikipedia~\cite{simplewiki}, and CommonLIT~\cite{commonlit},
among others, have developed texts at different levels of complexity for educational purposes.

The process of rewriting texts to different levels typically involves iterative editing
to ensure that the revised texts meet the desired difficulty criteria.
This readability assessment is based on various linguistic features,
with \textit{sentence length} and \textit{word frequency} identified as
key factors in previous studies~\cite{fitzgerald2015important}.
Although this process appears straightforward, accurately adjusting these elements
to achieve the target reading difficulty is challenging.
This task becomes even more complex for young learners,
where factors such as decodability~\cite{menton1999literature},
information load~\cite{landauer1997solution}, and other elements
play a significant role~\cite{fitzgerald2015important,fitzgerald2016examining}.

This paper introduces the \textbf{leveled-text generation} task.
As shown in \Cref{fig:task}, given (\textit{A}) a source text,
(\textit{B}) the readability level of the source text,
and (\textit{C}) the desired readability level,
the goal is to rewrite the given text so that the resulting text (\textit{i})
is within the desired readability level and (\textit{ii}) also preserves the original meaning.

As our first attempt at solving the leveled text generation task,
we assess the capability of three LLMs,
including \chatgpt~\cite{gpt35}, \llama~\cite{touvron2023llama}, and \mixtral~\cite{jiang2024mixtral},
to generate content at various levels of readability through zero-shot and few-shot prompting techniques.
Our evaluation involved 100 educational materials processed by \chatgpt, \llama, and \mixtral.
The findings indicate that providing a few examples will significantly improve performance in both manipulating the content readability and preserving the same information.
This might also suggest that the concept of the leveled-text generation and Lexile scale does not inherently exist,
as showing examples could help the model understand the task and realize what the texts should look like within the desired difficulty level.
When comparing different LLMs, we found that \llama was the most effective in achieving the desired difficulty range,
while \chatgpt excelled in maintaining the meaning of the original text.

We also conducted a manual quality inspection of 10 articles considering potential concerns
if the generated texts were used as learning materials.
Our analysis revealed a few issues:
(\textit{i}) the potential for introducing misinformation, particularly in the form of altered quotations and factual inaccuracies,
and (\textit{ii})
the inconsistent distribution of edits within the text, leading to an uneven readability level throughout the article.
These findings highlight the need for further research and development to address these concerns and
ensure the quality of the generated educational content.
This preliminary survey helped us identify five critical points for future advances in leveled text generation tasks.

\section{Related Work}
This task is related to (\textit{i}) text readability manipulation and (\textit{ii}) educational content generation.

\paragraph{Text Readability Manipulation.}
Most of the tasks related to manipulating text readability focus on text simplification.
\citet{S2009Feasibility} simplified texts through (i) rule-based conversion to sentence structures and (ii) replacing difficult words with easier synonyms.
\citet{Bingel2016Text} introduced a structured approach to text simplification using conditional random fields on dependency graphs to predict compressions and paraphrases.
\citet{Swain2019Lexical} developed an efficient text simplification technique using the WordNet model in NLTK.
\citet{Alkaldi2023Text} trained their own readability classifier and designed a reinforcement learning framework to train a text simplification model based on a GRU sequence-to-sequence model with attention. \citet{Alkaldi2023Text}'s work, however, is currently only trained in sentence-level simplification.
Recent advancements in LLMs have shown promising results.
\citet{Feng2023Sentence} investigated zero-shot and few-shot learning capabilities of LLMs, demonstrating superior performance.
\citet{Maddela2021Controllable} introduced a hybrid approach combining linguistically-motivated rules with a neural paraphrasing model.
Other studies focused on improving transparency and explainability.
\citet{Cristina2020Empirical} proposed a structured pipeline, analyzing text complexity prediction and complex component identification.

While these studies have made significant contributions,
further research is needed on generating texts at specific readability levels (including both increasing and decreasing text readability) while preserving meaning.
Our work on the leveled-text generation task benchmarks the LLMs performance and explores the possibility of this new topic.

\paragraph{Educational Content Generation.}
Recent studies have explored the potential of large language models (LLMs) in generating educational content. \citet{Leiker2023Prototyping} investigated the use of LLMs for creating adult learning content at scale;
\citet{MacNeil2022Automatically} focused on automatically generating computer science learning materials;
\citet{Gao2023Investigation} specifically investigated the application of LLMs to spoken language learning;
\citet{Jury2024Evaluating} evaluated LLM-generated worked examples in an introductory programming course;
and \citet{Xiao2023Evaluating} applied LLMs to generate reading comprehension exercises.
Despite the adoption of LLMs in various fields, this particular work focuses on leveraging their potential for language learning by rewriting texts to different readability levels.

\section{Benchmark the Leveled-Text Generation Task}

To establish a first benchmark for the leveled-text generation task,
we compiled a parallel dataset comprising 30K pairs of leveled texts and experimented with it using three LLMs,
namely \chatgpt, \llama, and \mixtral.
We evaluate the performance based on two aspects: (i) manipulating readability and (ii) content preservation.

Note that we chose the Lexile scale as a tool to assess readability,
given its wide adoption in educational settings to measure text complexity.
The Lexile Framework evaluates a student's reading skills and
the complexity of reading materials on the same scale~\cite{lexilereport}.
This makes it easier for students to choose books that are just right for their reading level,
helping them to enhance their language skills more effectively.
Given its wide adoption in K-12 educational settings to measure text complexity,
we decided that the Lexile measure was the best fit for our readability analysis.

\subsection{Task Definition}
As shown in \Cref{fig:task}, given (A) a source text,
(B) a readability level of the source text,
and (C) an intended readability level of the target text,
the leveled-text generation task should rewrite the source text to meet characteristics
within the intended readability level, such as the vocabulary used,
sentence structure, sentence length, and so on.


The Lexile Framework suggests that readability measures should rely on different features
for different groups of students.
For early-level learners (usually lower than grade 4 or a Lexile level of 750L),
readability relies on many specific features, such as decodability~\cite{menton1999literature},
information load~\cite{landauer1997solution}, phonetic features, word structure, sentence complexity, and so on~\cite{fitzgerald2015important,fitzgerald2016examining}.
For upper-level learners, the readability measurement mostly relies on
word frequency and sentence length~\cite{fitzgerald2015important}.
In this task, we would like to see if LLMs can learn the characteristics of the target Lexile level
and rewrite the source text accordingly.

\subsection{Evaluation Metrics.}
\kenneth{Again, results of what? I'm not clear what's the goal of this study.}
\kenneth{I assume this 100 sample will be reviewed by humans?}

We evaluate two aspects: (\textit{i}) whether the model could correctly rewrite the texts to the intended Lexile score and (\textit{ii}) whether the model could still preserve the same information.

To evaluate whether the resulting texts are aligned with the intended Lexile score,
we first used the Lexile Analyzer to measure the resulting texts and obtained the resulting Lexile score.
Several metrics were then calculated based on the intended and the resulting Lexile score:

\begin{enumerate}
\item \textbf{Mean Absolute Error (MAE)}, representing the absolute deviation between the intended and the resulting Lexile scores;
\item \textbf{Match Rate}, indicating the proportion of instances where the resulting Lexile score was within a $\pm$50 range of the intended score; and
\item \textbf{Directional Accuracy}, reflecting the proportion of instances where the resulting Lexile score moved in the intended direction (toward easier or more difficult levels).
\end{enumerate}

To measure information preservation, we used \textbf{BERTScores}~\cite{Zhang2020BERTScore}\footnote{We used the \texttt{microsoft/deberta-xlarge-mnli} model to obtain the embedding.},
semantic similarity~\cite{reimers-2019-sentence-bert}\footnote{We used sentence transformer with the \texttt{sentence-transformers/all-mpnet-base-v2} model},
and normalized edit distance 
to assess content preservation between the source texts and the resulting texts.

\subsection{Dataset}
We started with a leveled-text corpus, a collection of articles from various language-learning books organized into 1,690 sets.
Each set contains articles that share the same title, meaning they cover the same topic but are written at different readability levels to suit various reading abilities.
These articles range from two to six versions per set, with an average of 825 words per article.
The collection can be represented as follows:
\begin{equation}
    \text{Leveled Text Corpus} = \{\{A^1_1, A^1_2\}, \{A^2_1, A^2_2, ... A^2_6\}, ..., \{A^n_1, ..., A^n_m\}\}
\end{equation}
In this representation, $A^i_j$ refers to a specific language learning article, where $i$ indicates the set index and $j$ is the article index within that set.
Articles in the same group ($i$) discuss the same topic but at different complexity levels.
We assigned a readability score to each article by running the Lexile analyzer~\cite{lexilereport}.

The dataset was then split into three parts according to the set index: 90\% for training (1521 sets), 5\% for validation (84 sets), and 5\% for testing (85 sets).
To form a parallel dataset for the leveled-text generation task, we permuted all pairs of articles within each set.
This new parallel dataset consists of the following fields:
\begin{enumerate}
\item \textbf{Source text}: The original article that needs to be rewritten to match a different readability level.
\item \textbf{Source Lexile score}: The readability score of the source text, used to guide the adjustment process in terms of simplifying or complicating the text.
\item \textbf{Target text}: The rewritten article, adjusted to the desired readability level.
\item \textbf{Target Lexile score}: The intended readability score for the rewritten article, serving as a goal for the leveled-text generation task.
\end{enumerate}

After preprocessing, the train, validation, and test sets comprised 29,990, 1680, and 1700 pairs of leveled texts, respectively.
Note that although our test set comprised 1700 instances, we only selected 100 samples for this preliminary analysis.

\subsection{Leveled-Text Generation with LLMs}
As the first attempt to solve the leveled-text generation task,
we experimented with prompting popular LLMs, namely \chatgpt, \llama, and \mixtral.
The purpose of this method is to explore the capabilities of LLMs in generating leveled texts
and to establish a baseline for future improvements.

We tried prompting techniques with both zero-shot learning and few-shot learning.
For zero-shot learning, we defined the Lexile score and then provided the source text, source Lexile score, and target Lexile score to the model (see \Cref{appendix:prompt} for the actual prompt used.)

However, the definition of Lexile score can still be vague for LLMs.
To address this, we also tried few-shot learning, where actual examples from the training set were presented
to teach LLMs what the text within a particular readability level should look like.

The few-shot learning examples for each sample were chosen based on the corresponding Lexile scores of the source and target texts.
We identified training samples whose source and target Lexile scores were both within a 50-point range of the target sample's corresponding scores:
\begin{equation}
    \begin{aligned}
        &\text{A training sample is qualified if:} \\
        &\begin{Bmatrix}
            |Lexile_{train-source} - Lexile_{sample-source}| \leq 50 \\
            |Lexile_{train-target} - Lexile_{sample-target}| \leq 50
        \end{Bmatrix}
    \end{aligned}
\end{equation}

From the qualifying training samples, we selected the \textbf{shortest} $n$ samples to minimize the context size of the prompt.
In the few-shot prompt, we included the source text, source Lexile score, target text, and target Lexile score as examples for in-context learning (see \Cref{appendix:prompt} for the actual prompt used).


\begin{table*}[]
\addtolength{\tabcolsep}{-.8mm}
\begin{tabular}{@{}llcccccccccc@{}}
\toprule
\multirow{2}{*}{\textbf{Method}} & \multirow{2}{*}{\textbf{Model}} & \multirow{2}{*}{\textbf{\#Shot}} & \multirow{2}{*}{\textbf{Support}} & \multicolumn{3}{c}{\textbf{Lexile Score}} & \multicolumn{3}{c}{\textbf{BERTScore}} & \multirow{2}{*}{\makecell{\textbf{Semantic} \\ \textbf{Similarity}$\uparrow$}} & \multirow{2}{*}{\makecell{\textbf{Normalized} \\ \textbf{Edit Distance}}} \\ \cmidrule(lr){5-7} \cmidrule(lr){8-10}
 &  &  &  & \textbf{MAE}$\downarrow$ & \textbf{Match}$\uparrow$ & \textbf{Direction}$\uparrow$ & \textbf{Precision}$\uparrow$ & \textbf{Recall}$\uparrow$ & \textbf{F1}$\uparrow$ & & \\ \midrule
\multirow{3}{*}{\textbf{Zero-shot}} & \textbf{GPT-3.5} & 0 & 100 & 257.6 & 15.00\% & 80.00\% & 79.87\% & 76.54\% & 78.03\% & 0.893 & 0.941 \\
 & \textbf{LLaMA-2 70B} & 0 & 100 & 206.5 & 15.15\% & 71.72\% & 75.49\% & 71.82\% & 73.56\% & 0.894 & 0.947 \\
 & \textbf{Mixtral 8x7B} & 0 & 100 & 256.0 & 11.00\% & 79.00\% & 74.69\% & 73.74\% & 74.18\% & 0.894 & 0.951 \\ \midrule
\multirow{3}{*}{\textbf{Few-shot}} & \textbf{GPT-3.5} & 3 & 100 & 205.3 & 15.00\% & 75.00\% & 82.85\% & 80.18\% & 81.45\% & 0.937 & 0.934 \\
 & \textbf{LLaMA-2 70B} & 1 & 99 & 172.9 & 22.22\% & 86.87\% & 72.96\% & 71.25\% & 73.01\% & 0.887 & 0.949 \\
 & \textbf{Mixtral 8x7B} & 3 & 100 & 210.9 & 12.00\% & 83.00\% & 72.89\% & 69.88\% & 71.27\% & 0.935 & 0.929 \\ \bottomrule
\end{tabular}
\addtolength{\tabcolsep}{+.8mm}
\caption{Performance comparison for different models. In Lexile score measurements, MAE refers to the mean absolute error between the resulted Lexile score and the intented Lexile score; Match measures whether the resulted Lexile scores fall within the range of the intented Lexile score $\pm$ 50; and Direction measures whether the resulted Lexile score moves toward the intended direction (\ie, easier or harder).}
\label{tab:llm-performance}
\end{table*}

\subsection{Benchmark Results}
\Cref{tab:llm-performance} shows the benchmark results. 
Note that for few-shot learning, we tried 1-shot, 3-shot, and 5-shot samples,
but only the best-performing one was reported.

Our findings suggest that providing a few examples significantly improves performance,
as few-shot learning outperforms zero-shot learning.
This phenomenon also indicates that the concept of ``leveled-text generation''
with respect to the Lexile scale does not inherently exist in these LLMs.
To fully utilize LLMs, it is necessary to teach them what Lexile means and
what the texts would possibly look like at each level.

When comparing different LLMs, we found that \llama performs best in adjusting readability,
whereas \chatgpt demonstrates superior performance in preserving content and meaning.
However, a major concern is that although \llama achieves the best Lexile Score measure
(lowest MAE = 172.9; highest match rate = 22.22\%),
its normalized edit distance is also the highest,
meaning that it produces the least edit to the content
(see \Cref{sec:discussion}: \textit{Bias in Current Models}
for the discussion of the desired edit behavior).
Another concern when using \llama is its smaller context size (4K tokens),
which can cause the task to fail if the prompt exceeds the context size limitation.

\begin{figure*}
    \centering
    \begin{subfigure}{0.33\textwidth}
        \centering
        \includegraphics[width=1.0\linewidth]{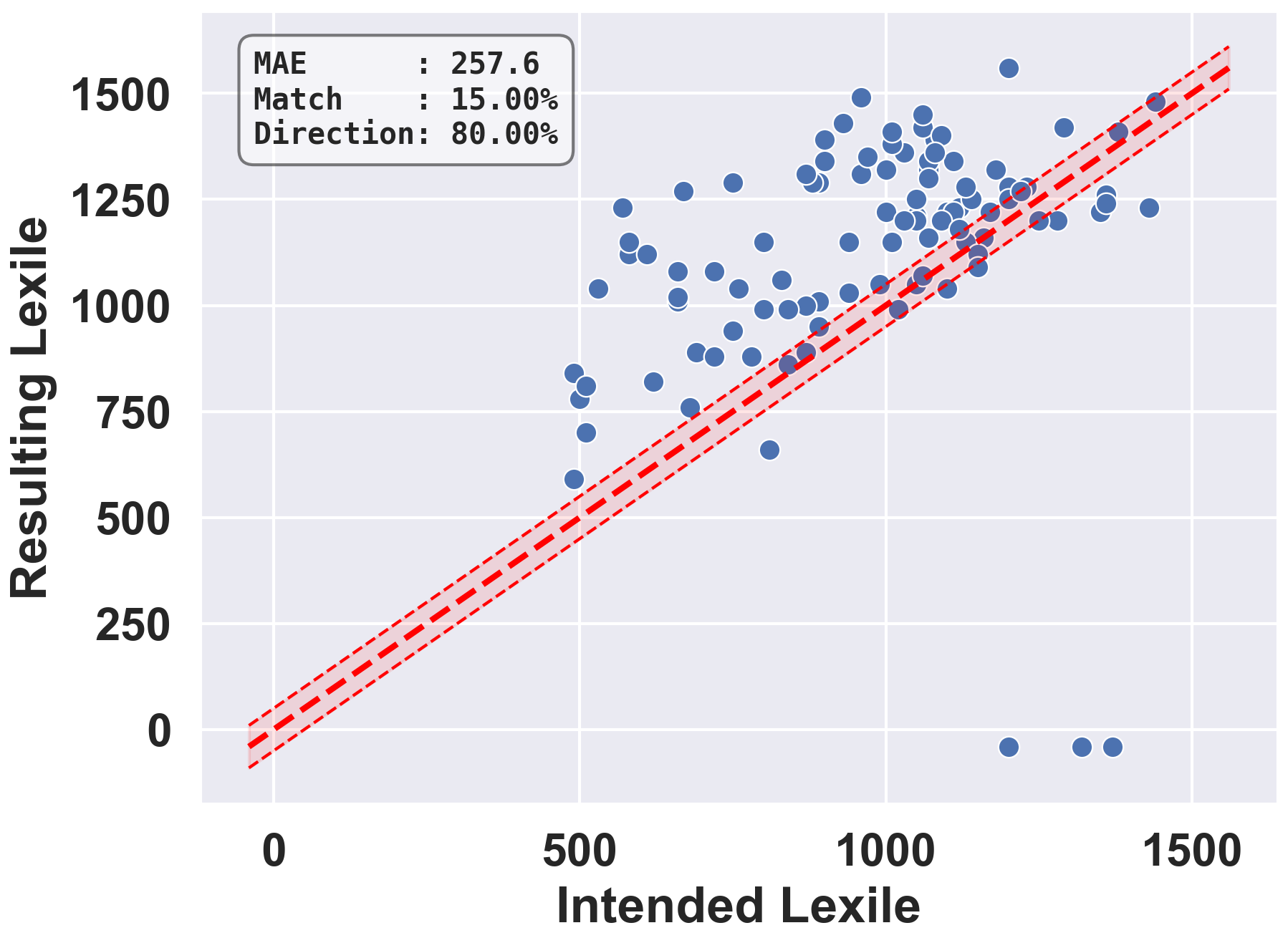}
        \caption{Zero-shot \chatgpt}
    \end{subfigure}
    \begin{subfigure}{0.33\textwidth}
        \centering
        \includegraphics[width=1.0\linewidth]{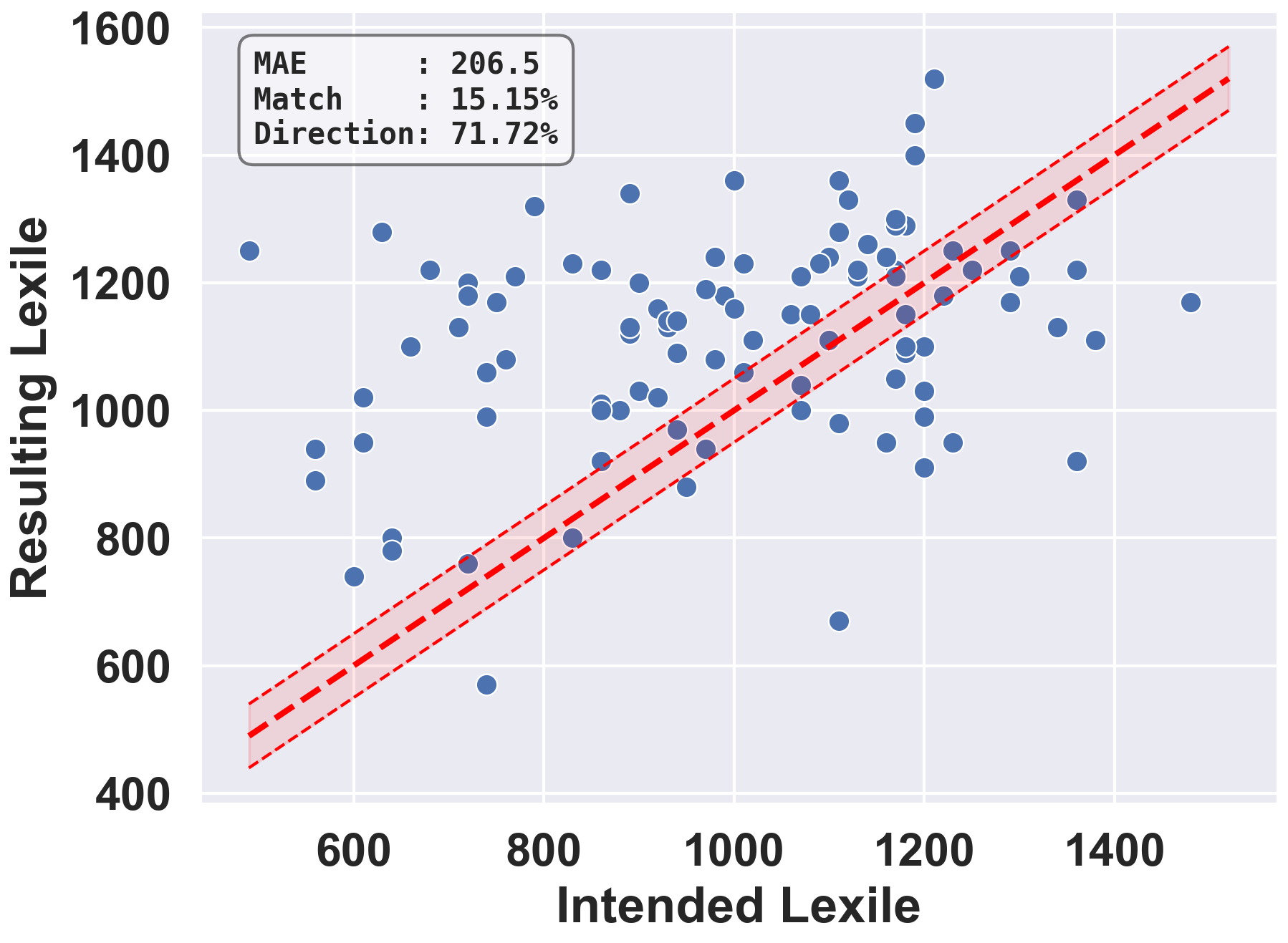}
        \caption{Zero-shot \llama}
    \end{subfigure}
    \begin{subfigure}{0.33\textwidth}
        \centering
        \includegraphics[width=1.0\linewidth]{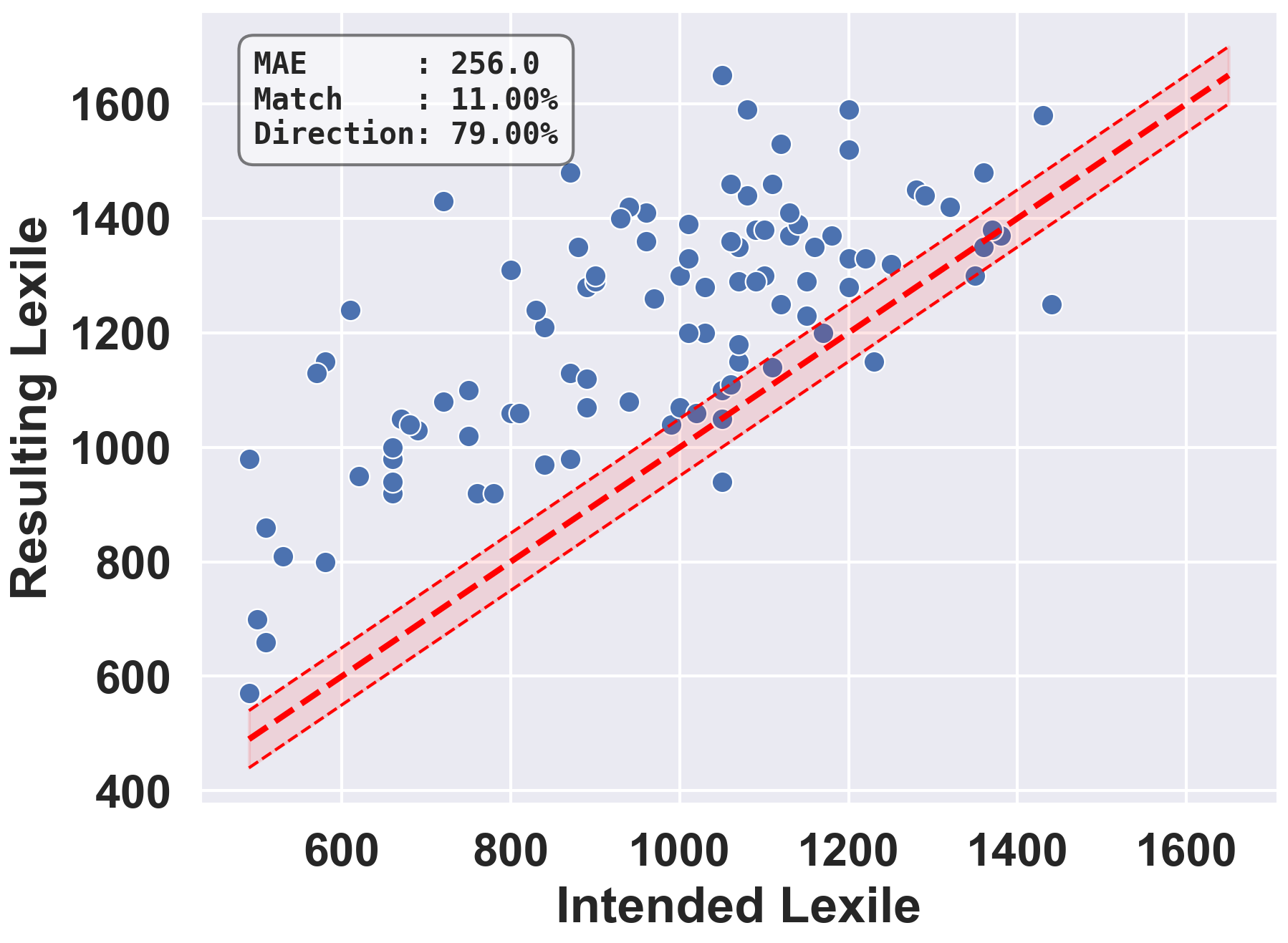}
        \caption{Zero-shot \mixtral}
    \end{subfigure}

    \begin{subfigure}{0.33\textwidth}
        \centering
        \includegraphics[width=1.0\linewidth]{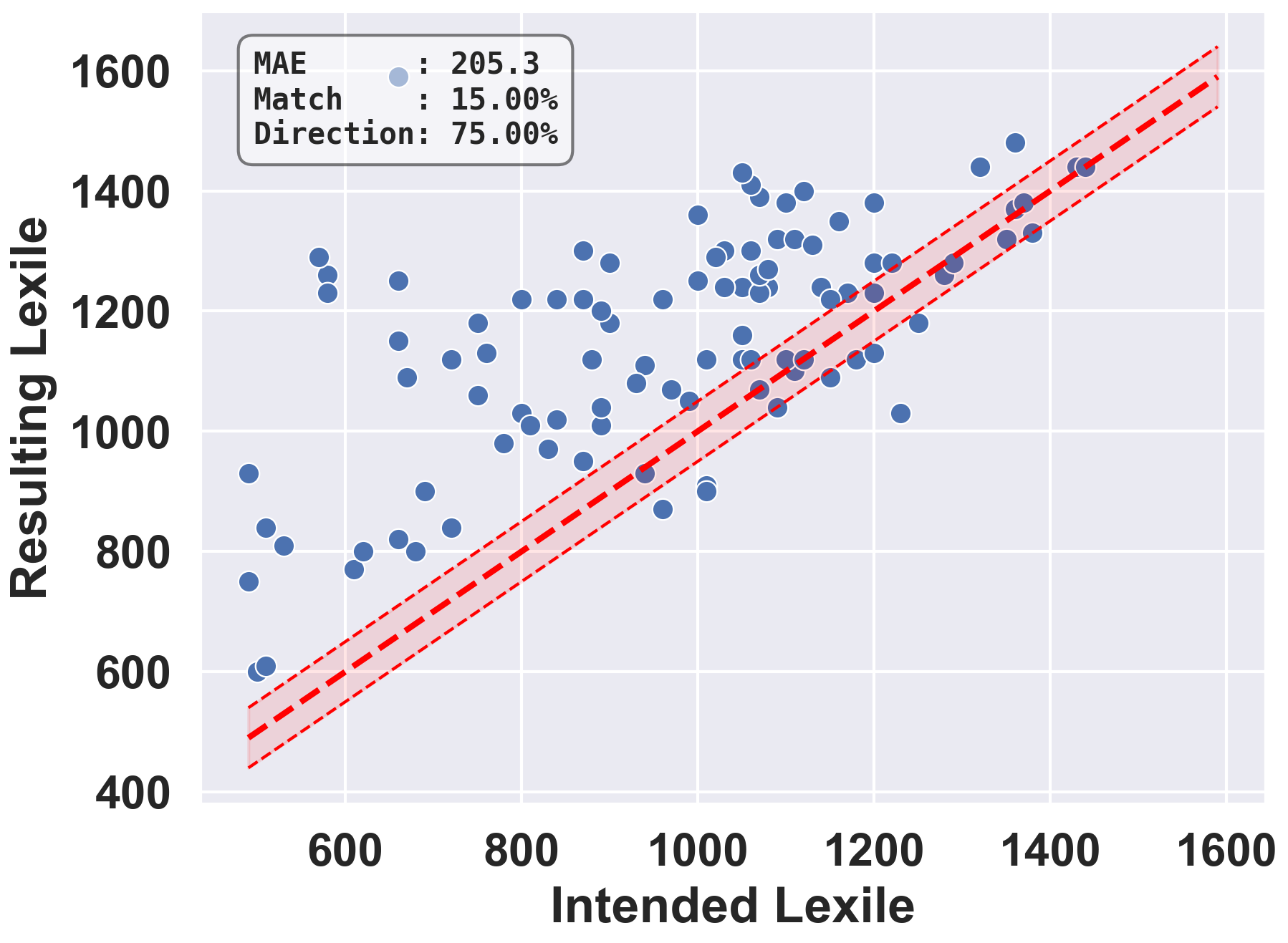}
        \caption{Few-shot \chatgpt}
    \end{subfigure}
    \begin{subfigure}{0.33\textwidth}
        \centering
        \includegraphics[width=1.0\linewidth]{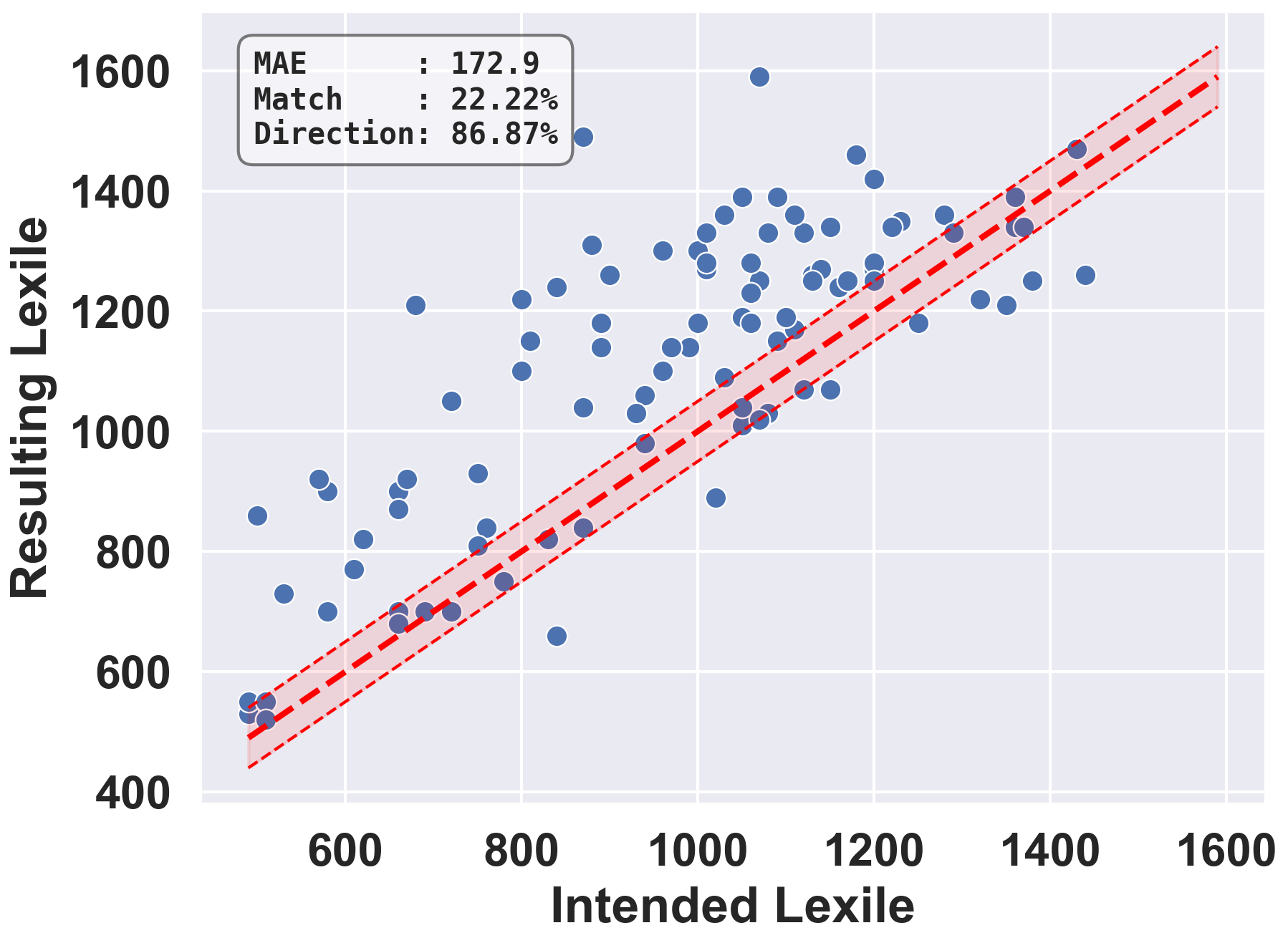}
        \caption{Few-shot \llama}
    \end{subfigure}
    \begin{subfigure}{0.33\textwidth}
        \centering
        \includegraphics[width=1.0\linewidth]{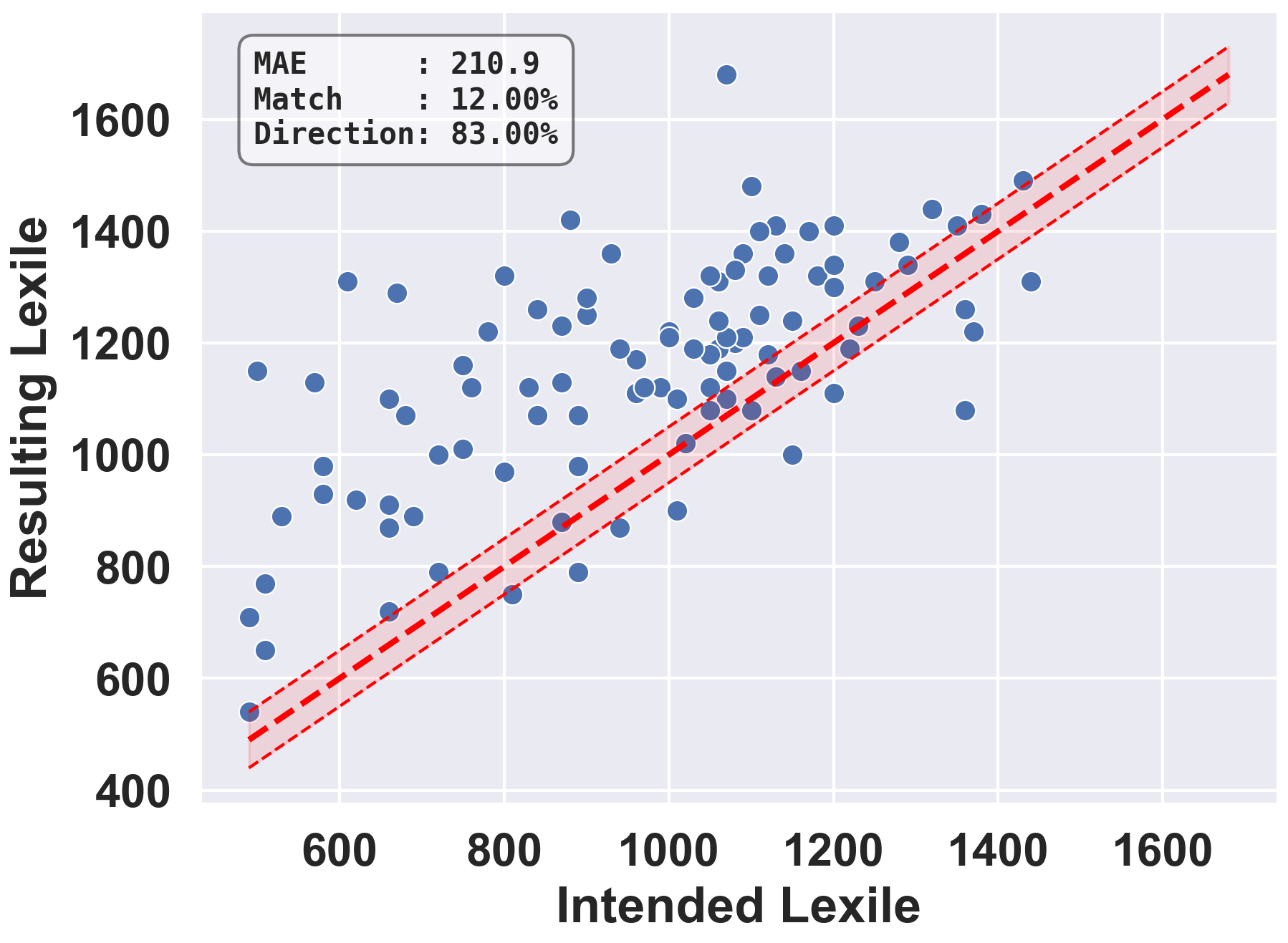}
        \caption{Few-shot \mixtral}
    \end{subfigure}

    \caption{Scatter plots comparing intended and resulting Lexile scores for text generated by GPT-3.5, LLaMA-2 70B, and Mixtral 8x7B models in zero-shot and few-shot settings. The red-shaded area represents the region where resulting scores are within $\pm$50 points of the intended scores. A higher proportion of data points fall above the red area, indicating that the resulting Lexile scores tend to skew higher than the intended scores, suggesting a tendency for the models to generate slightly more complex text than the target difficulty level, regardless of the specific model or prompting approach used.}
    \label{fig:score-distribution}
\end{figure*}

\begin{figure*}
    \centering
    \begin{subfigure}{0.33\textwidth}
        \centering
        \includegraphics[width=1.0\linewidth]{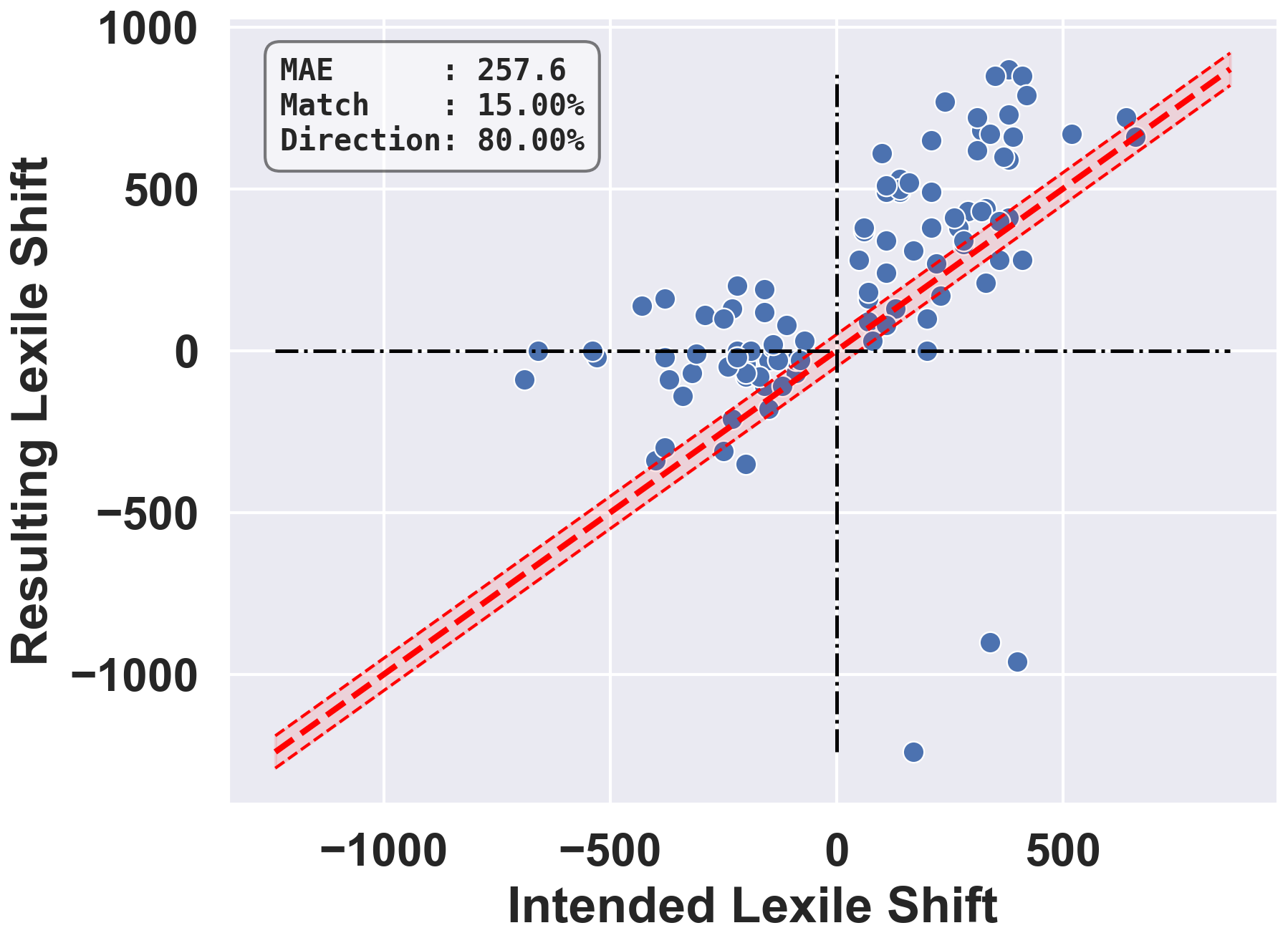}
        \caption{Zero-shot \chatgpt}
    \end{subfigure}
    \begin{subfigure}{0.33\textwidth}
        \centering
        \includegraphics[width=1.0\linewidth]{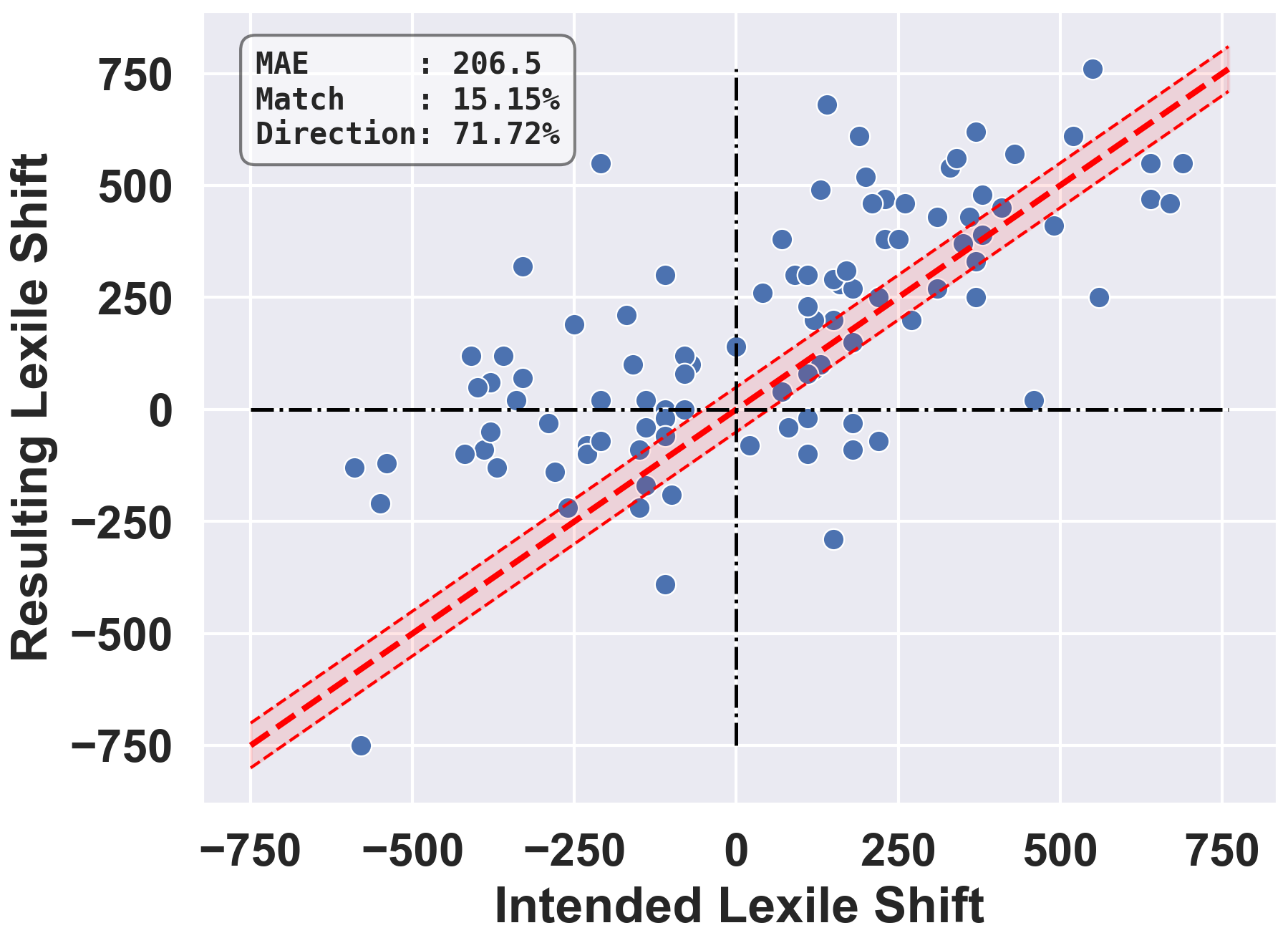}
        \caption{Zero-shot \llama}
    \end{subfigure}
    \begin{subfigure}{0.33\textwidth}
        \centering
        \includegraphics[width=1.0\linewidth]{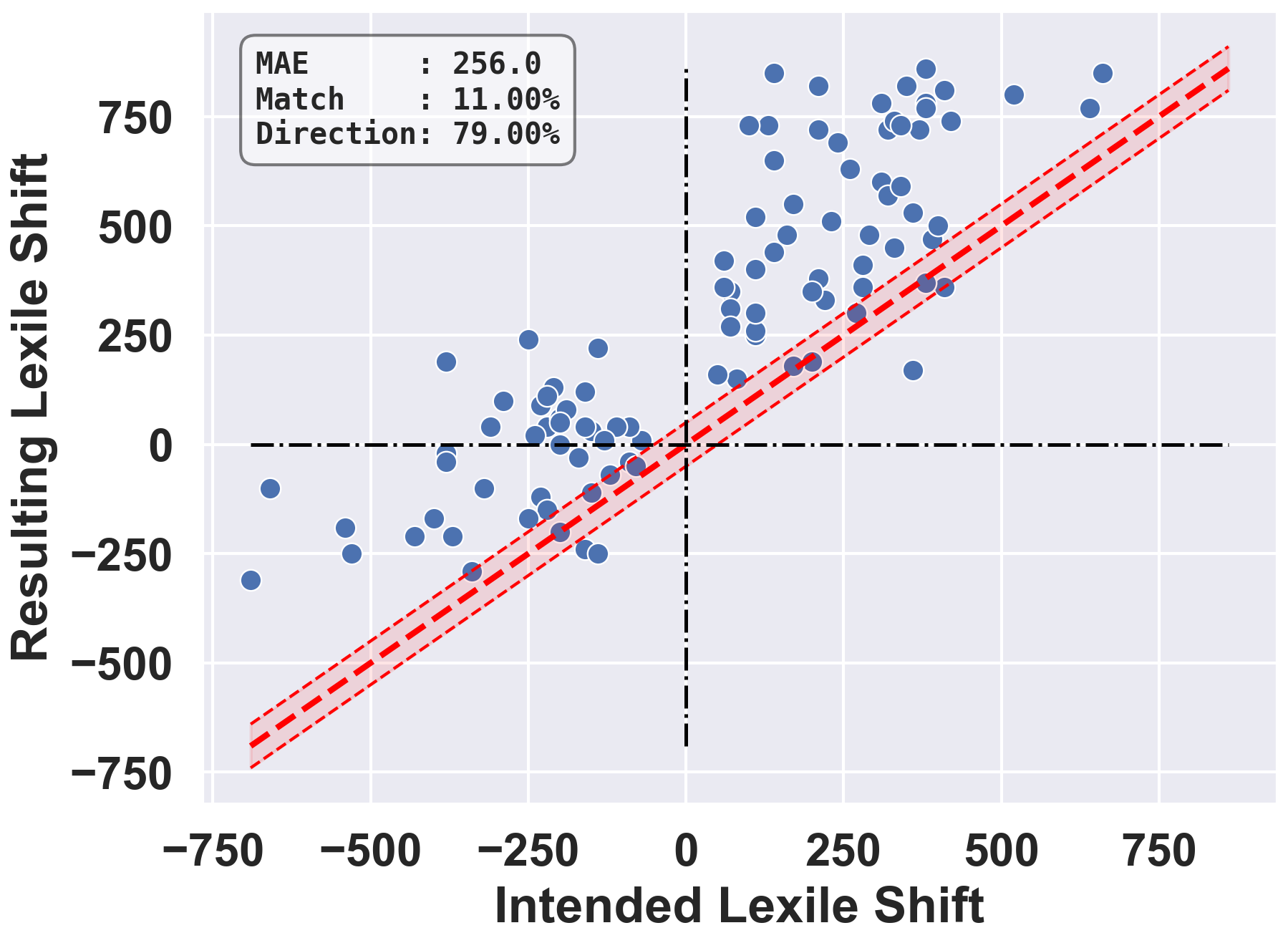}
        \caption{Zero-shot \mixtral}
    \end{subfigure}

    \begin{subfigure}{0.33\textwidth}
        \centering
        \includegraphics[width=1.0\linewidth]{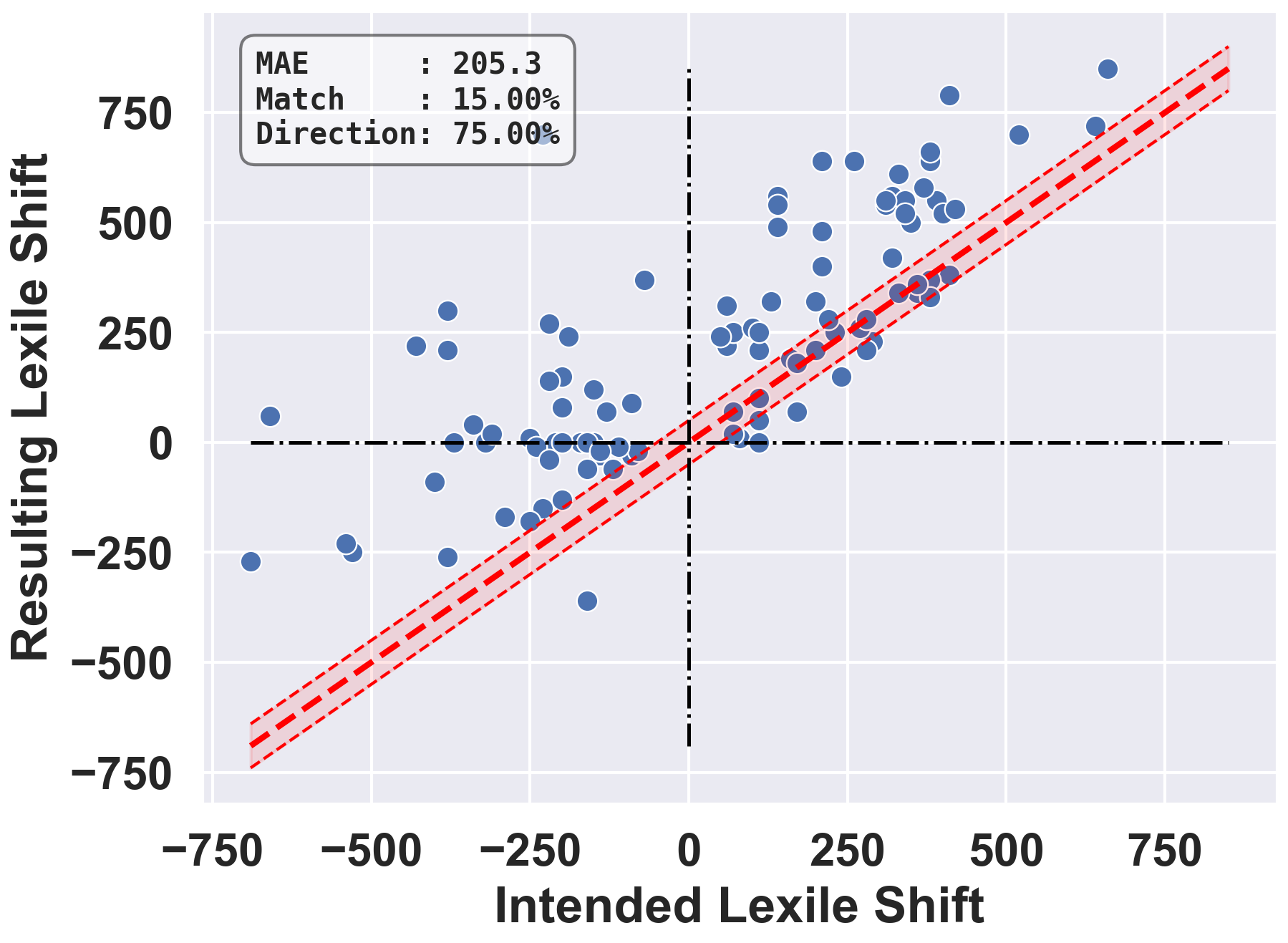}
        \caption{Few-shot \chatgpt}
    \end{subfigure}
    \begin{subfigure}{0.33\textwidth}
        \centering
        \includegraphics[width=1.0\linewidth]{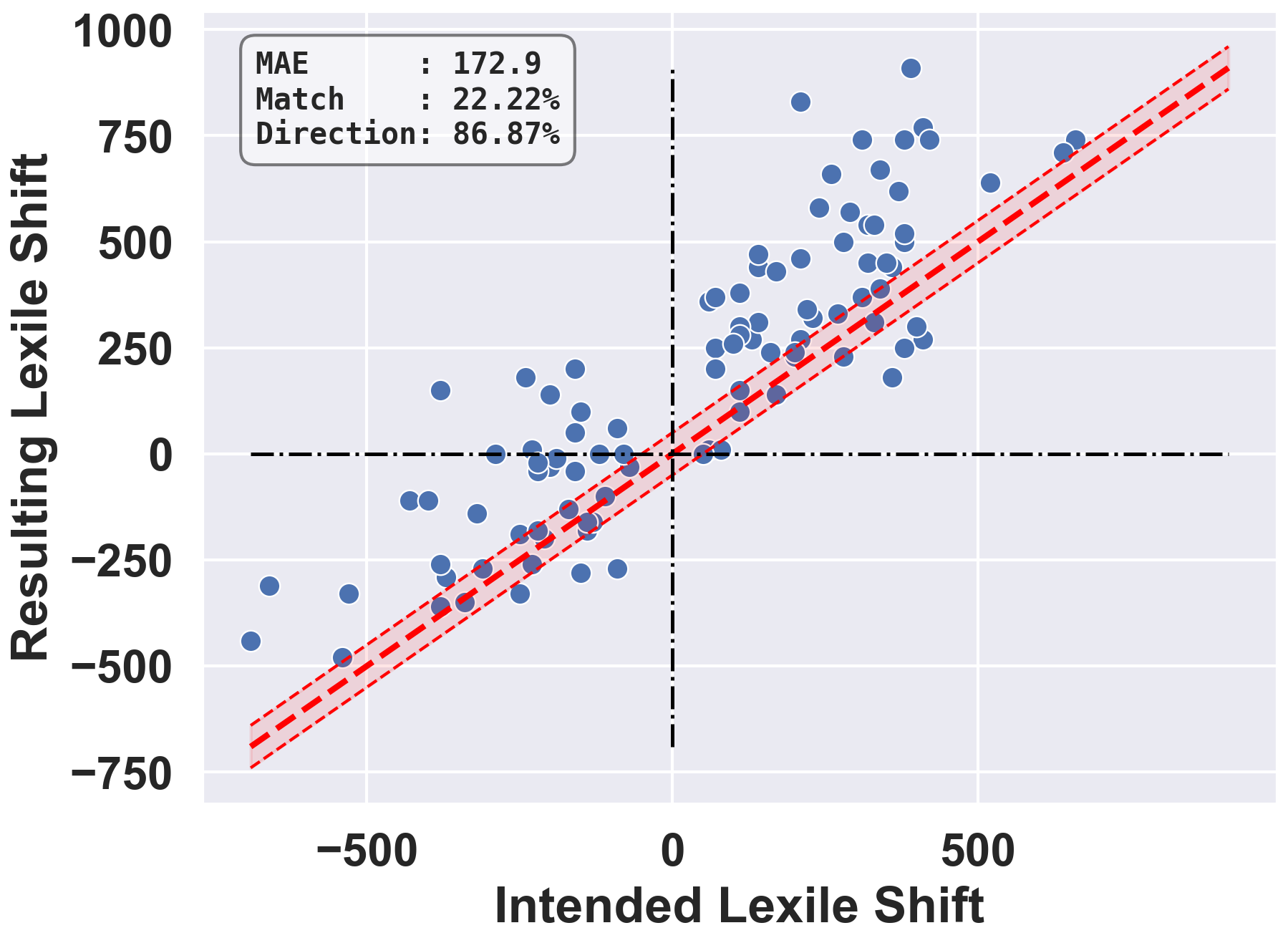}
        \caption{Few-shot \llama}
    \end{subfigure}
    \begin{subfigure}{0.33\textwidth}
        \centering
        \includegraphics[width=1.0\linewidth]{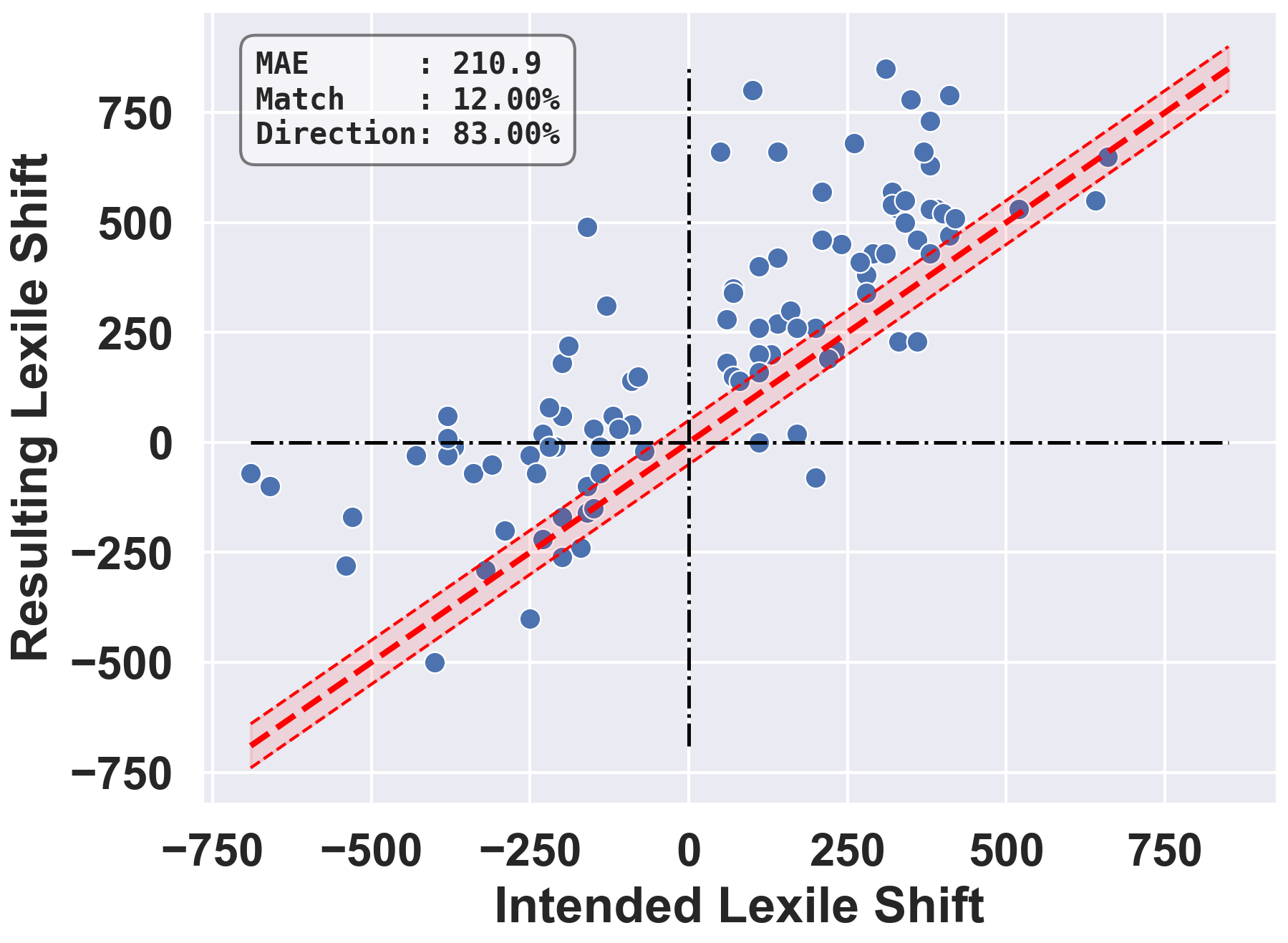}
        \caption{Few-shot \mixtral}
    \end{subfigure}

    \caption{Scatter plots comparing intended and resulting Lexile shifts for text generated by GPT-3.5, LLaMA-2 70B, and Mixtral 8x7B models in zero-shot and few-shot settings. The Lexile shift is calculated as the difference between the intended or resulting Lexile score and the source Lexile score. Data points falling within the first and third quadrants indicate the correct direction of change in text complexity. However, the overall distribution of points still exhibits a skew towards higher difficulty levels, suggesting that the models tend to generate text that is slightly more complex than the intended shift, regardless of the specific model or prompting approach employed.}
    \label{fig:shift-distribution}
\end{figure*}

In \Cref{fig:score-distribution} and \Cref{fig:shift-distribution}, we present scatter plots to illustrate the distribution of intended Lexile versus resulting Lexile scores and intended Lexile shift versus resulting Lexile shift.
These figures provide valuable insights into the behavior of the language models.
The red-shaded area represents the region where resulting scores fall within $\pm$50 points of the intended scores.
In both \Cref{fig:score-distribution} and \Cref{fig:shift-distribution}, we can clearly observe that a higher proportion of the data points are located \textbf{above} the red-shaded area, indicating that the generated texts are generally more complex than the intended level.
Although this bias has been identified, more research is necessary to understand its underlying causes and develop potential solutions to mitigate its effects.

\section{Detailed Inspection and Discussion}
\label{sec:discussion}

We conducted a manual investigation of 10 samples selected from the test set.
We reviewed the generated texts and discussed potential issues in using these outputs as educational materials.

\paragraph{Problems with Text Shortening and Lengthening}
Changing content readability may involve shortening or lengthening the given text.
LLMs are good at shortening texts with advanced prompting techniques such as Chain-of-Density~\cite{adams2023sparse}.
Expanding texts, however, requires the introduction of new information.
While this is less problematic for narrative genres, it poses significant challenges
for factual content like science and news,
where maintaining accuracy and minimizing misinformation are crucial.
In our review of 30 generated articles, we noticed several modifications to quotations,
which is not ideal.
Recent studies suggest that Retrieval-Augmented Generation (RAG)
can mitigate hallucinations~\cite{shuster-etal-2021-retrieval-augmentation},
but implementing an RAG system that integrates up-to-date, external information remains a challenge.

\paragraph{Limitations in Leveled-Text Generation}
Leveled-text generation, particularly for scientific materials,
might not be feasible for all Lexile levels,
especially for younger learners.
Paraphrasing sentences or selecting frequently appearing words have their limits in changing text difficulty.
For example, explaining ``photosynthesis'' to kids might require a new explanation
that involves age-appropriate analogies or visuals.
Thus, the leveled-text generation task, which aims to rewrite texts,
may not adequately address such needs.
To significantly change the original content to achieve the desired readability,
a more sophisticated LLM might be able to complete the task,
but it is outside the scope of the current leveled-text generation task.

\paragraph{Incorporating Educational Lessons}
In educational contexts, materials usually come with learning objectives,
such as grammar, vocabulary, knowledge, etc.
Integration of these educational elements into rewritten texts remains an unresolved challenge.
Moreover, determining the appropriate learning objectives for students
at different levels is crucial yet challenging.
We believe human involvement is essential, but more research is needed
to explore how such involvement can be done.

\paragraph{Important Information Should Remain Unchanged}
There is a need among content creators to retain specific pieces of information,
such as key terms (e.g., ``Photosynthesis''), essential sentences (e.g., a quote),
or particular sections deemed more important than others.
Current LLMs may address this requirement through prompt engineering.
However, developing an intuitive interface that allows users to
(\textit{i}) highlight areas of text that should remain unchanged and
(\textit{ii}) verify whether the generated texts meet these criteria is essential.

\paragraph{Limitations and Bias in Current Models}
We identified biases in the three LLMs.
First, we found a tendency for the models to produce shorter texts than the originals
(Original: 825 words, Few-Shot: 350-500 words),
regardless of whether the intention was to simplify or complexify the texts.
This bias may have been influenced by the use of shorter few-shot samples.
However, a similar pattern emerged in zero-shot scenarios
(Original: 825 words, Zero-Shot: 500-600 words),
suggesting a possible bias in off-the-shelf LLMs.

Second, the distribution of edits within articles often appeared uneven,
with some paragraphs remaining unchanged while others underwent significant revisions (this appears more frequently in the texts generated by \llama).
This inconsistent editing pattern is unsuitable for educational materials,
even if the articles achieve the desired readability.
More research is needed to identify the causes and develop solutions.

\section{Conclusion}
Our investigation into leveled-text generation using LLMs underscores the potential and challenges of automating educational content creation.
While \llama shows promise in adjusting text complexity, \chatgpt is better at maintaining content meaning.
Future research should address the nuanced demands of educational content,
including accurate representation of information,
integration of learning objectives,
and retention of key information,
to enhance leveled-text generation process.


\bibliographystyle{styles/ACM-Reference-Format}
\bibliography{bib/main}

\appendix

\section{Prompt}
\label{appendix:prompt}
Here, we provide the prompt we used for the experiments.

The \textbf{zero-shot prompt} we used is as follow. We quickly introduce the Lexile measurement and then provide 
(\textit{i}) \{SOURCE-TEXT\}, (\textit{ii}) \{SOURCE-LEXILE\}, and (\textit{iii}) \{TARGET-LEXILE\} as the information.

\begin{prompt}
A Lexile measure is defined as ``the numeric representation of an individual's reading ability or a text's readability (or difficulty),''
where lower scores reflect easier readability and higher scores indicate harder readability.\\

In this task, we are trying to rewrite a given text into the target Lexile level and keep the original meaning and information.
Given the original draft (Lexile = \{SOURCE-LEXILE\}):\\

[TEXT START]

\{SOURCE-TEXT\}

[TEXT END] \\

Rewrite the above text and \{TASK\} to the difficulty level of Lexile = \{TARGET-LEXILE\}.
\end{prompt}

The \textbf{few-shot prompt} we used is as follow.
There are three sections in the prompt, introduction, example, and task (separated by ``START-INTRO'' tags).
The section tag will not appear in the final prompt.
For few-shot learning with more than 1 shot, only the example section will be repeated several times.
\begin{prompt}
\textbf{((START-INTRO))} \\
A Lexile measure is defined as "the numeric representation of an individual's reading ability or a text's readability (or difficulty),"
where lower scores reflect easier readability and higher scores indicate harder readability. \\

In this task, we are trying to rewrite a given text into the target Lexile level and keep the original meaning and information. \\
\textbf{((END-INTRO))} \\

\textbf{((START-EXAMPLE))} \\
Here is an example. \\
Lexile = \{SOURCE-LEXILE\}

[TEXT START]

\{SOURCE-TEXT\}

[TEXT END] \\

Rewritten Text of Lexile = \{TARGET-LEXILE\}

[TEXT START]

\{TARGET-TEXT\}

[TEXT END]

\textbf{((END-EXAMPLE))} \\

\textbf{((START-TASK))} \\
Now, given the original text (Lexile = \{SOURCE-LEXILE\}):

[TEXT START]

\{SOURCE-TEXT\}

[TEXT END] \\

Rewrite the above text and \{TASK\} to the difficulty level of Lexile = \{TARGET-LEXILE\}.
Do not include [TEXT START] and [TEXT END] in your response. Thanks. \\
\textbf{((END-TASK))} \\
\end{prompt}

\end{document}